\documentclass[letterpaper, 10 pt, conference]{ieeeconf}
\usepackage{times}
\usepackage{graphicx}
\usepackage{amsmath,amssymb,amsopn,amstext,amsfonts}
\usepackage{cancel}
\usepackage[space]{cite}
\usepackage{pdfsync}
\usepackage{balance}
\usepackage{color}
\usepackage{mathtools}
\usepackage{algpseudocode}
\usepackage{algorithm} %\usepackage[ruled,vlined,linesnumbered]{algorithm2e}
\usepackage{bm}

\usepackage{diagbox}
\usepackage{float}
\usepackage{epstopdf}
\usepackage{pifont}
\usepackage{fixltx2e}
\usepackage{amsmath}
\usepackage{multirow}
\usepackage{url}
\usepackage{verbatim}
\usepackage{float}
\usepackage{stfloats}
\usepackage{placeins}
\makeatletter
\let\NAT@parse\undefined
\makeatother
\usepackage[linkcolor=black,citecolor=black,urlcolor=black,colorlinks=TRUE]{hyperref}
\usepackage{booktabs}
\usepackage{graphicx}
\usepackage{subcaption}

\graphicspath{{./}}
\DeclareGraphicsExtensions{.png,.jpg,.eps,.pdf}
\IEEEoverridecommandlockouts
\makeatletter
\def\endthebibliography{
	\def\@noitemerr{\@latex@warning{Empty `thebibliography' environment}}
	\endlist
}

	\title{\LARGE \bf CREPES: Cooperative RElative Pose Estimation System}
	
	\author{Zhiren Xun\textsuperscript{$\dagger$ 1,2}, Jian Huang\textsuperscript{$\dagger$ 1,2}, Zhehan Li\textsuperscript{1,2},
     \\Zhenjun Ying\textsuperscript{2}, Yingjian Wang\textsuperscript{1,2}, Chao Xu\textsuperscript{1,2}, Fei Gao\textsuperscript{1,2}, and Yanjun Cao\textsuperscript{1,2}
		\thanks{\textsuperscript{$\dagger$} \textbf{Equal contribution.}}
		\thanks{This work was supported by National Nature Science Foundation of China under Grant 62103368.(Corresponding author: Yanjun Cao, Fei Gao.)}
		\thanks{
		  \textsuperscript{1} State Key Laboratory of Industrial Control Technology, Institute of Cyber-Systems and Control, Zhejiang University, Hangzhou, 310027, China.
                }
        \thanks{ \textsuperscript{2} Huzhou Institute of Zhejiang University, Huzhou, 313000, China.}
        \thanks{E-mails:\tt\small \{xzr, huangjian2022, zhehanli, yj\_wang, cxu, fgaoaa, yanjunhi\}@zju.edu.cn, yzj@stu.csust.edu.cn}
	}

\begin{document}
	\makeatletter
	\let\@oldmaketitle\@maketitle
	\renewcommand{\@maketitle}{\@oldmaketitle
		\includegraphics[width=\linewidth,height=0.32\linewidth]{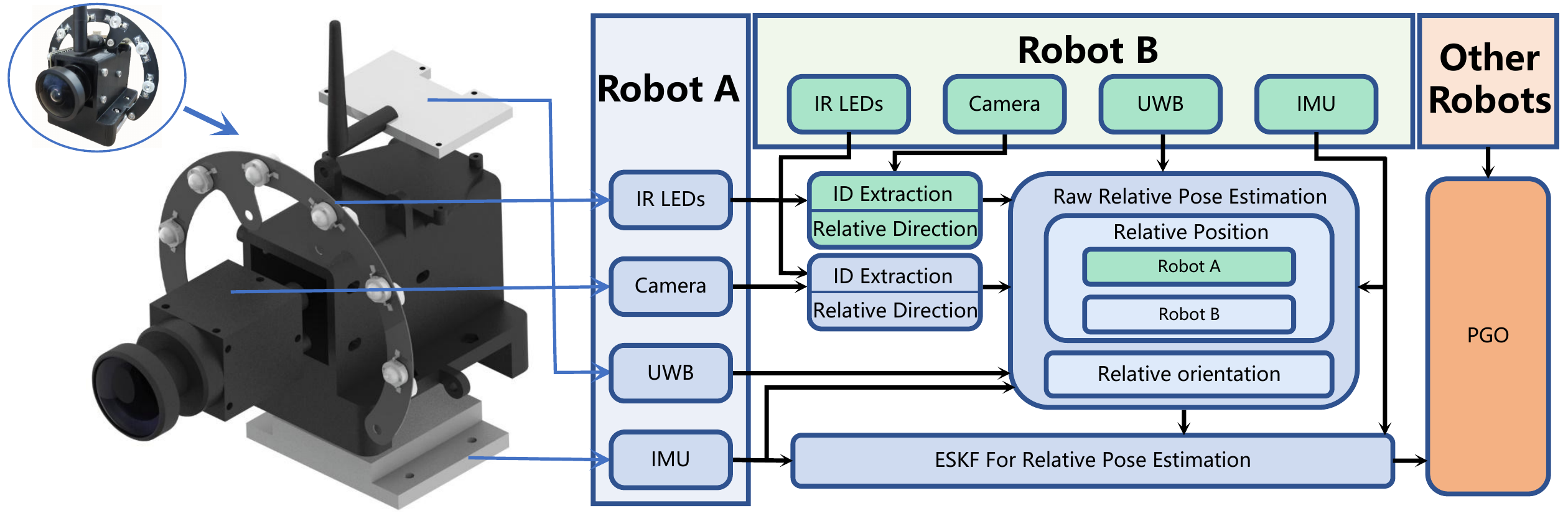}
		\captionsetup{font={small}}
		\captionof{figure}{	\label{pic:toutu}
			Hardware design and software architecture of CREPES. The hardware consists of an \textit{IMU}, an \textit{UWB}, \textit{IR LEDs} and a \textit{Camera}. The \textit{Raw Relative Pose Estimation} software module collects neighbors' ID (from \textit{ID Extraction}), directional measurement (from \textit{Relative Direction}), distance (from UWB) and IMU information to get the raw estimation. Then an \textit{ESKF} module filters the result using the IMU information. When multiple neighbors are around, a \textit{PGO} module is employed to further improve the performance.
		}
		\vspace{-0.25cm}	
	}
	\makeatother
	\maketitle
 	\setcounter{figure}{1}	
	\begin{abstract}
	
	Mutual localization plays a crucial role in multi-robot cooperation. 
	%In this work, we propose a novel system to estimate the 3D relative pose targeting real-world applications. 
	CREPES, a novel system that focuses on six degrees of freedom (DOF) relative pose estimation for multi-robot systems, is proposed in this paper. 
	CREPES has a compact hardware design using active infrared (IR) LEDs, an IR fish-eye camera, an ultra-wideband (UWB) module and an inertial measurement unit (IMU). By leveraging IR light communication, the system solves data association between visual detection and UWB ranging. Ranging measurements from the UWB and directional information from the camera offer relative 3-DOF position estimation. Combining the mutual relative position with neighbors and the gravity constraints provided by IMUs, we can estimate the 6-DOF relative pose from a single frame of sensor measurements. In addition, we design an estimator based on the error-state Kalman filter (ESKF) to enhance system accuracy and robustness. When multiple neighbors are available, a Pose Graph Optimization (PGO) algorithm is applied to further improve system accuracy. We conduct enormous experiments to demonstrate CREPES' accuracy between robot pairs and a team of robots, as well as performance under challenging conditions.
	
	%We conduct experiments in various environments, and the results show that our system outperforms state-of-the-art accuracy and robustness, especially in challenging environments.

	\end{abstract}

	\section{Introduction}
	\label{sec:introduction}

	Recently, multi-robot systems have received increasing attention due to their high efficiency in many fields, such as collaborative mapping,   exploration\cite{https://doi.org/10.48550/arxiv.2109.07764}, monitoring\cite{lin2018topology} and search and rescue\cite{sherman2018cooperative}. For an efficient multi-robot system, mutual relative localization is the key to accomplishing tasks cooperatively. Stable, accurate and fast relative pose estimation between robots can significantly improve the quality of collaboration. For instance, robots can continuously transform neighbors' perceptions into their frames to acquire a robust collaborative perception. 
	
	% One common practice for relative localization is using robots' odometry in a global reference frame, like satellite-based global positioning system (GPS)\cite{4724195}, motion capture system (MCS)\cite{7989376} and UWB system with multiple anchors\cite{7353810}. The relative pose can be calculated from 6-DOF subtraction between agents' global states. However, these systems rely on pre-installed infrastructure or require time-consuming calibration, and not applicable to robots in unknown environments. Simultaneous localization and mapping (SLAM) can provide each robot with the odometry in its map frame. Relative transformations between multiple robots can be estimated by matching common features in their maps, either centralized or distributed. Nevertheless, they usually need high computational resources and communication bandwidth, which is challenging to scale. Having direct robot-to-robot observations can produce instant relative pose estimation. As can be seen in many systems where robots are equipped with specially designed structures, such as AprilTags \cite{5979561} and LEDs \cite{8967660}. However, short detection range, strict viewpoint requirements and sensitivity to ambient light limit their application in multi-robot systems.
 
    One common practice for relative localization is using robots' odometry in a global reference frame, like satellite-based global positioning system (GPS)\cite{4724195}, motion capture system (MCS)\cite{7989376} and UWB system with multiple anchors\cite{7353810}. The relative poses can be calculated from the subtraction between agents' global states. However, these systems rely on pre-installed infrastructure or require time-consuming calibration, and not applicable to robots in unknown environments. Simultaneous localization and mapping (SLAM) can provide each robot with the odometry in its global reference frame. Relative transformations between multiple robots can be estimated by matching common features in their maps, either centralized or distributed. Nevertheless, they usually need high computational resources and communication bandwidth. By equipping robots with specially designed structures, such as AprilTags \cite{5979561} and LEDs \cite{8967660}, the relative pose can be estimated from direct robot-to-robot  observations in many systems. However, short detection range, strict viewpoint requirements and sensitivity to ambient light limit their application in multi-robot systems.

    To overcome difficulties of dependence on the infrastructure and environment, high computational cost and low adaptability, we design a Cooperative RElative Pose Estimation System (CREPES) for multi-robot systems. CREPES, which can obtain instant 6-DOF relative poses to all the neighbors in a large-scale environment, consists of a novel hardware design and supported software. The hardware consists of active IR LEDs, an IR fisheye camera, an IMU and an UWB.
    The system can produce a raw estimation of the relative pose between two robots from one single frame of mutual observations.
    To cope with multiple sources of sensor noise, we establish a relative motion model and apply an adapted ESKF \cite{sola2017quaternion}  where the reference frame is in motion. A pose graph optimization strategy is applied when two or more neighbors are around to further improves the accuracy.
    In summary, our contributions are as follows:
    \begin{itemize}
		\item [1)] 
		We propose CREPES, a novel relative pose estimation system that produces accurate relative position and orientation within one-shot mutual observations.
		\item [2)]
		We design and implement the hardware prototype consisting of active IR LEDs, an IR fisheye camera, an IMU and an UWB.
		\item [3)]
		We design a relative pose estimator based on ESKF, where the reference frame is in motion.
		\item [4)]
		We propose a PGO-based algorithm to improve the accuracy when multiple neighbors are around.

	\end{itemize}	
	
	\section{Related Work}
	\label{sec:related_works}
   While many infrastructure-based systems (e.g. GPS, MCS, UWB with anchors\cite{7989376,nguyen2016ultra}) circumvent the mutual localization by using global poses, we focus on real-time relative pose estimation for autonomous navigation in an unknown environment. 
   We classify current systems into direct and indirect methods depending on whether the relative pose can be estimated instantaneously.  
    \subsection{Indirect Methods}

    Multi-robot SLAM is a typical indirect method in which agents estimate the relative transformation between robots' map frames by matching common features in their maps, either in a centralized or distributed fashion. Centralized works \cite{schmuck2019ccm}, \cite{8456343}, \cite{8360045} usually require a powerful central server to collect keyframes from all agents and optimize their trajectories through the global bundle adjustment in a common coordinate frame. The relative information between agents can be acquired from the server directly. Distributed methods \cite{lajoie2020door}, \cite{9662965} rely on inter-robot loops to estimate relative poses between robots' coordinate frames in a distributed manner. In these works, robots need to exchange map feature descriptors for inter-robot loop detection. More importantly, the feature descriptors should be generated from similar viewpoints to improve accuracy. 
    
	Mutual observations, such as relative ranging or bearing, are applied to help reduce the high dependency on the environment and inter-loop detection.
    Cao \cite{cao2021vir} proposes an efficient method by combining the visual inertial odometry (VIO) system with mutual UWB ranging measurements between robots and an anchor. Wang \cite{https://doi.org/10.48550/arxiv.2203.09312} utilizes trajectories of Unmanned Aerial Vehicles (UAVs) and anonymous bearing measurements to formulate mutual localization as a mixed-integer quadratically constrained quadratic problem and obtain a certifiably global optimum.
    Xu \cite{DBLP:journals/corr/abs-2103-04131} fuses omnidirectional visual inertial SLAM and UWB measurements with global graph-based optimization.      
    Overall, these methods may suffer from degeneration due to the reliance on SLAM systems in the feature-less environment.
    
    Researchers also explored ranging/bearing only systems to further reduce environmental dependency. Zhou \cite{zhou2008robot} provides theoretical proof that the minimum number of distance constraints required for 3-DOF relative pose estimation is five. Guo \cite{guo2017ultra} proposes an infrastructure-free cooperative 3-DOF relative localization system with UWB measurements and applies it to real-world UAV formation control. Trawny \cite{4399075} puts forth an algebraic algorithm using ten range measurements to estimate 6-DOF relative pose. However, these methods usually require enough motion excitation over long trajectories in practice.
    
    \subsection{Direct Methods}
    
    Although direct methods typically need customized hardware, the self-sufficiency, stability, efficiency and accuracy still attract enormous attention.
    Cutler \cite{cutler2013lightweight} proposes a lightweight solution for estimating ranging and bearing relative to a known marker, which consists of three IR LEDs in a fixed pattern. Faessler \cite{faessler2014monocular} utilizes four infrared LEDs structures following certain rules and uses the Perspective-n-Point (PnP) algorithm to calculate the relative pose between a quadrotor and a ground robot.
    When coming to the multi-robot scenario, active markers\cite{dias2016board} or active LEDs coded board\cite{8967660} is designed to encode ID information either by pulsating capabilities or LEDs arrangements. However, since the utilization of the PnP algorithm, these methods usually work at a short distance to keep the LED light spots distinguishable in the image. By using Ultraviolet LEDs and estimating the bearing vector and distance, Walter \cite{walter2018fast} has dramatically improved the detected range, with a maximum working distance of 15 meters.
  
    UWB is getting popular in multi-robot systems due to its low cost and good ranging accuracy.
	Fishberg \cite{fishberg2022multi} provides an inter-agent 3-DOF relative pose estimation system for robots in a 2D plane, where each agent is equipped with four UWB modules. The relative pose is calculated by modeling observed ranging biases and systematic antenna obstructions in a nonlinear least squares optimization. 
	Cossette \cite{cossette2022optimal}  presents a method for computing optimal formations for relative pose estimation, during which both the relative position and relative heading of the agents with two UWB modules are locally observable. An on-manifold gradient descent procedure is used to determine optimal formations for improving estimation. Since the noise property of UWB, the baseline distance between multiple UWB modules should be far to acquire good performance, which limits the platform size to use these systems.

    \section{Relative Pose Estimation System}
    
     Our novel system includes a compact hardware design and supported software. As shown in Fig. \ref{pic:toutu}, the hardware system includes IR LEDs, an IR fisheye camera, an IMU, and an UWB, and the software contains the ID Extraction module, raw relative pose estimation module, ESKF filter module, and PGO module. ID Extraction module establishes the data association between ranging measurement and directional information by using the IR camera and a disc-shaped IR LED board. The raw relative pose estimation module gets direct relative position and orientation. Then, we design a relative movement model and adapt ESKF to filter raw estimations. For systems with more than two robots, the PGO module launches to further improve accuracy.

    \subsection{Hardware}	 
	Fig. \ref{pic:toutu} shows an overview of sensor components and their physical settings. A disc-shaped board with six 950nm IR LEDs is designed to transmit ID information. We program an ARM Cortex-M3 STM32 microcontroller to control the flickering of LEDs for ID encoding. Correspondingly, we use a MV-SUA133GM camera made by MindVision, equipped with a 950 nm IR filter, to decode the ID information. The camera has a fisheye lens with a 185 degrees field of view (FOV) and is set to a frame rate of 200 $\mathbf{Hz}$ (maximum 245 $\mathbf{Hz}$), with a global shutter. We use a DW1000-based UWB module from NoopLoop to provide mutual ranging. It uses a dongle antenna to get relatively good omnidirectional ranging and a maximum range of 500 meters, with a standard deviation of 5 centimeters.
	%Distance measurement accuracy of 10cm with a standard deviation of 5cm in the x- and y-axis direction and 30cm with a standard deviation of 15cm in the z-axis direction.
	%Distance measurement accuracy of approx. 10 cm in x- and y-axis and approx. 30 cm in z-axis.
	In addition, a 6-DOF low-cost MEMS IMU module is used to provide accelerations and angular velocities at a frequency of 100 $\mathbf{Hz}$. 
	The accelerometer noise density is $183.3 \mu g/ \sqrt\mathbf{Hz}$ and the gyroscope noise density is $0.021 ^{\circ}/s / \sqrt\mathbf{Hz}$. 
	In practice, we use imu-tk \cite{tpm_icra2014} to perform calibration to correct imprecise scaling factors and axes misalignments. Lastly, we use an Intel NUC with i5 processor as the computation platform and its onboard WiFi as the communication medium. Note that we configure the WiFi network card into a self-organizing MESH mode (BATMAN network \footnote{https://www.open-mesh.org/projects/open-mesh/wiki/BATMANConcept}) to remove the dependence on a central router.

    \subsection{ID Extraction}
    The ID information for each node is encoded into the LED pulsating control. 
    IR LED boards are programmed with designed duty rates using a 50ms period. Benefitting from the infrared filter, all lighted IR LED boards are easily distinguished from natural features in a captured image. First, we convert the image into binary with a threshold and perform circle detection using Hough transform to obtain pixel coordinates of the centers of detected spots. Next, the detected spots are associated with previous ones according to a distance constraint, and the duty rates are calculated. Finally, IDs are decided by comparing the calculated duty rates with elements in an ID library. At the same time, the pixel coordinates of the detected spots' centers work as the directional measurement for pose estimation. 
 
    \subsection{Raw Relative Pose Estimation}
    \label{sec:direct computation}
    	\begin{figure}[h]
    		\vspace{0.5cm}
    		\centering
    		\includegraphics[width=0.95\linewidth]{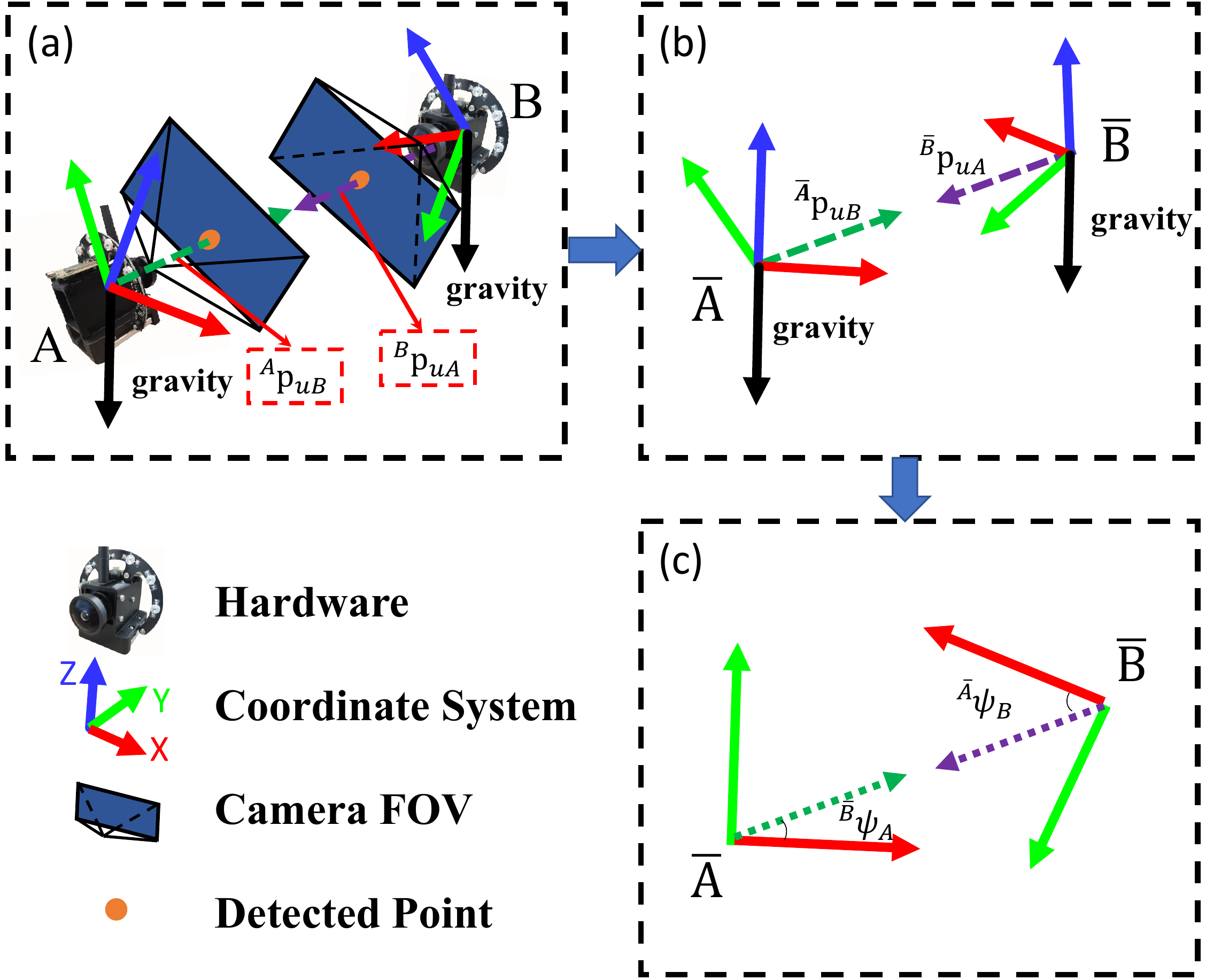}
    		\captionsetup{font={small}}
    		\caption{
    			The pipeline of the raw relative pose estimation algorithm. (a) We use the DS fisheye camera reprojection model to compute two-unit directional vectors $^{{B}}\mathbf{p}_{uA}$ and $^{{A}}\mathbf{p}_{uB}$. (b) By using the gravity alignment, we get the expressions $^{\bar{B}}\mathbf{p}_{uA}$ and $^{\bar{A}}\mathbf{p}_{uB}$ of the two-unit directional vectors under the two intermediate frames $\bar{A}$ and $\bar{B}$. (c) we project $^{\bar{B}}\mathbf{p}_{uA}$ and $^{\bar{A}}\mathbf{p}_{uB}$ onto the x-o-y plane and compute the angles between each projected vector and the corresponding x-axis. 
    		}
    		\label{pic:Direction Computation}
    		\vspace{-0.6cm}
    	\end{figure}
    The raw relative pose estimation uses mutual directional measurement from the camera (described in the above section), UWB ranging, own IMU and neighbor's IMU measurements, as shown in Fig. \ref{pic:toutu}. 
    The working pipeline of how we estimate raw mutual relative pose is shown in Fig. \ref{pic:Direction Computation}, taking robots A and B as an example (B as the observer).  Firstly, when two robots observe each other in their image frame, they extract ID and pixel coordinates from the ID extraction module. Based on the Double Sphere (DS) projection model of the fisheye camera \cite{usenko2018double},
    we get the unit directional vector $^B\mathbf{p}_{uA} \in \mathbf{R}^3$ for detected robot A in robot B's frame. 
    The relative position of robot A in B's frame ${^B}\mathbf{\widetilde{p}}_A$ can be calculated by
    \begin{equation}
        ^B\mathbf{\widetilde{p}}_A = d_{AB} * {^B}\mathbf{p}_{uA}
    \end{equation}
    where $d_{AB}$ is the ranging measurement from UWB between robots A and B.
    
    To estimate the relative orientation, we introduce intermediate frames ${\bar{A}}$ and ${\bar{B}}$.   
    We use $Z-Y-X$ Euler angles to define roll, pitch and yaw, and extract roll and pitch angles using the gravity constraint. As shown in Fig. \ref{pic:Direction Computation} (b), by rotating the roll and pitch angles to align the z-axis of robot B's body frame opposite gravity's direction, we get the new ${\bar{B}}$ frame. The unit directional vector $^B\mathbf{p}_{uA}$ in the ${\bar{B}}$ frame is expressed as $^{\bar{B}}\mathbf{p}_{uA}$, where
    \begin{equation}
        ^{\bar{B}}\mathbf{p}_{uA} = \mathbf{R}_{pitch}^B \mathbf{R}_{roll}^B  {^B}\mathbf{p}_{uA}
    \end{equation}
    $\mathbf{R}_{pitch}^B$ and $\mathbf{R}_{roll}^B$ are obtained from robot B's IMU measurements. As shown in Fig. \ref{pic:Direction Computation} (c), we project $ ^{\bar{B}}\mathbf{p}_{uA}$ to the X-O-Y plane to get an angle $^{\bar{B}}{\psi}_A$ between the projected vector and the positive direction of the x-axis. In the same way, we take robot A as the observer and can also get $^{\bar{A}}{\psi}_B$. Therefore the relative yaw angle $\psi$ can be defined as,
    \begin{equation}
        \psi = {^{\bar{B}}{\psi}_A} - {^{\bar{A}}{\psi}_B} + \pi
    \end{equation}
    After all, we calculate the relative orientation matrix $^B{\mathbf{\widetilde{R}}_A}$ by
    \begin{equation}
    %    ^B{\mathbf{\widetilde{R}}_A} = {{\mathbf{R}_{roll}^B}^{T}} {{\mathbf{R}_{pitch}^B}^{T}} \mathbf{R}_{yaw} \left\{ \psi \right\} {\mathbf{R}_{roll}^A} {\mathbf{R}_{pitch}^A}
	^B{\mathbf{\widetilde{R}}_A} = {{\mathbf{R}_{roll}^B}^{T}} {{\mathbf{R}_{pitch}^B}^{T}} \mathbf{R}_{yaw} \left\{ \psi \right\} {\mathbf{R}_{pitch}^A} {\mathbf{R}_{roll}^A}
    \end{equation}
    where $\mathbf{R}_{yaw}\left\{ \psi \right\}$ is the rotation matrix corresponding to $\psi$, $\mathbf{R}_{pitch}^A$ and $\mathbf{R}_{roll}^A$ are extracted from robot A's IMU measurements.
    
    Similarly, by taking robot A as the observer, we can also calculate the relative position $^A\mathbf{\widetilde{p}}_B$ and relative orientation $^A{\mathbf{\widetilde{R}}_B}$ in robot A's frame.
    
    \subsection{ESKF Filter}
    \label{sec:ESKF section}
    To improve the estimation quality, we adapt an ESKF to filter the raw relative pose estimations. Compared to the typical state estimation in a normal inertial system, our ESKF model takes extra consideration of the reference frame motion. Same as the above section, we keep robot B as the observer. 
        
    \subsubsection{Prediction Model}
    In a static reference frame $W$, we define ${^{{W}}}{\mathbf{q}}_{(\cdot)}$, ${^{{W}}}{\mathbf{p}}_{(\cdot)}$ and ${^{{W}}}\mathbf{v}_{(\cdot)}$ as the quaternion, position and velocity of the robot $(\cdot)$, respectively. For robots A and B, the relative state can be calculated by
    \begin{subequations}
    \begin{align}
	   	^B\mathbf{p}_A &= \mathbf{R}^T \left\{ {^{{W}}\mathbf{q}}_B \right\}({^{ W}}{\mathbf{p}}_{A} - {^{W}}{\mathbf{p}}_{B}) \label{motion equations in w a} \\
	   	^B\mathbf{v}_A &= \mathbf{R}^T \left\{ {^{W}\mathbf{q}}_B \right\}({^{ W}}{\mathbf{v}}_{A} - {^{W}}{\mathbf{v}}_{B}) \label{motion equations in w b}\\
	   	^B\mathbf{q}_A &= {^{W}\mathbf{q}}_B^* \otimes {^{W}\mathbf{q}}_A \label{motion equations in w c}
    \end{align}
    \end{subequations}
    where $\mathbf{R} \left\{ \mathbf{q} \right\}$ and $\mathbf{R} \left\{ \mathbf{\bm{\theta}} \right\}$ are the rotation matrices associated with the quaternion $\mathbf{q}$ and the angular vector $\bm{\theta}$, respectively, $\mathbf{R}^T\left\{ \cdot \right\}$ is the inverse matrix of $\mathbf{R}\left\{ \cdot \right\}$ and  $\otimes$ represents the quaternion product. It should be noted that $^B\mathbf{v}_A$ is not the time rate of the change of $^B\mathbf{p}_A$ and the relationship is revealed in equation (\ref{nominal_state_start}).
    
    For simplicity, we write $^B\mathbf{p}_A$, $^B\mathbf{v}_A$, and $^B\mathbf{q}_A$ as $\mathbf{p}$, $\mathbf{v}$, and $\mathbf{q}$, respectively.  
    We define that $\mathbf{x}$ is the nominal state, $\mathbf{x}_t$ is the true state and ${\delta}\mathbf{x}$ is the error state,
    \begin{equation}
        \mathbf{x}=\left[\begin{array}{c}
                \mathbf{p} \\
                \mathbf{v} \\
                \mathbf{q}
            \end{array}\right]\qquad 
        \mathbf{x}_t=\left[\begin{array}{c}
                \mathbf{p}_t \\
                \mathbf{v}_t \\
                \mathbf{q}_t
            \end{array}\right]\qquad 
        \mathbf{{\delta}x}=\left[\begin{array}{c}
                \mathbf{{\delta}p} \\
                \mathbf{{\delta}v} \\
               {\delta}\bm{\theta}_A\\
               {\delta}\bm{\theta}_B
            \end{array}\right]
    \end{equation}
    where ${\delta}\bm{\theta}_{(\cdot)}$ is the small local angular error used to parameterize an error quaternion of the robot $(\cdot)$, ${\delta}\mathbf{q}_{(\cdot)}\approx\left[\begin{array}{c}1\\\frac{{\delta}\bm{\theta}_{(\cdot)}}{2}\end{array}\right]$. The true state can be computed with nominal-state and error-state by
    \begin{align}
        \mathbf{x}_t &= \mathbf{x} \oplus \delta \mathbf{x} \label{true_state_equation} \\
        \mathbf{p}_t&=\mathbf{R}^T\left\{{\delta \bm{\theta}}_B\right\}(\mathbf{p}+{\delta}\mathbf{p}) \tag{\ref{true_state_equation}a} \\
        \mathbf{v}_t&=\mathbf{R}^T\left\{{\delta \bm{\theta}}_B\right\}(\mathbf{v}+{\delta}\mathbf{v}) \tag{\ref{true_state_equation}b} \\
        \mathbf{q}_t&={{\delta}\mathbf{q}}_B^*{\otimes}\mathbf{q}{\otimes}{{\delta}\mathbf{q}}_A \tag{\ref{true_state_equation}c} 
    \end{align}
    where ${{\delta}\mathbf{q}}_B^*$ is the conjugate quaternion of ${{\delta}\mathbf{q}}_B$. 
    
    We take the robot ${(\cdot)}$ IMU acceleration measurements $\mathbf{a}_{m(.)}$ and gyroscope measurements $\bm{w}_{m(\cdot)}$  as the ESKF filter input $\mathbf{u}_m$. The input noise vector $\mathbf{u}_n$ consists of acceleration noise $\mathbf{a}_{n(.)}$ and gyroscope noise $\bm{w}_{n(\cdot)}$. $\mathbf{a}_{n(\cdot)}$ and $\bm{w}_{n(\cdot)}$ are modeled by white Gaussian processes. 
    % The input noisys $n_i$ , which are modeled by white Gaussian processes, consist of acceleration noisy $a_{n(.)}$ and gyroscope noisy $w_{n(.)}$.Their mean is zero.
    \begin{equation}
        \mathbf{u}_m=\left[\begin{array}{c}
                \mathbf{a}_{mA} \\
                \bm{w}_{mA} \\
                \mathbf{a}_{mB} \\
                \bm{w}_{mB}
            \end{array}\right]\qquad 
        \mathbf{u}_n=\left[\begin{array}{c}
                \mathbf{a}_{nA} \\
                \bm{w}_{nA} \\
                \mathbf{a}_{nB} \\
                \bm{w}_{nB}
            \end{array}\right]\qquad 
    \end{equation}
    
    We have the system model of the nominal-state as
    \begin{subequations}
    \begin{align}
        \mathbf{p} &\gets \mathbf{R}^T\left\{\bm{w}_{mB}\Delta t \right\}(\mathbf{p} + 
        \mathbf{v} \Delta t + \frac{1}{2}(\mathbf{R}\left\{\mathbf{q}\right\} \mathbf{a}_{mA} 
        - \mathbf{a}_{mB}){\Delta t}^2) \label{nominal_state_start}  \\
        \mathbf{v} &\gets \mathbf{R}^T\left\{\bm{w}_{mB}\Delta t \right\} (\mathbf{v} + (\mathbf{R}\left\{\mathbf{q}\right\} \mathbf{a}_{mA} - \mathbf{a}_{mB}) \Delta t)  \\
        \mathbf{q} &\gets \mathbf{q}^*\left\{ \bm{w}_{mB} \Delta t \right\} \otimes \mathbf{q} \otimes \mathbf{q} \left\{ \bm{w}_{mA} \Delta t \right\} \label{nominal_state_end} 
    \end{align}
    \end{subequations}
	where $\gets$ stands for a discrete time update, $\Delta t$ is the discrete time interval and $\mathbf{q} \left\{ \bm{\theta} \right\}$ is the quaternion associated with the angular vector $\bm{\theta}$.
	
    We write the differential equations of the error-state as
    \begin{align}
        \delta \mathbf{x} &\gets f(\mathbf{x}, \delta \mathbf{x}, \mathbf{u}_m, \mathbf{u}_n)=\mathbf{F_x}(\mathbf{x}, \mathbf{u}_m) \delta \mathbf{x} + \mathbf{F_i} \mathbf{u}_n \label{delta_x} \\
        \delta \mathbf{p} &\gets \mathbf{R}^T\left\{\bm{w}_{mB} \Delta t \right\}(\delta \mathbf{p} + \delta \mathbf{v} \Delta t) \tag{\ref{delta_x}a} \label{delta_x_a}\\
        \delta \mathbf{v} &\gets \mathbf{R}^T\left\{\bm{w}_{mB} \Delta t \right\}(\delta \mathbf{v} + \bm{\alpha} \Delta t) \tag{\ref{delta_x}b} \\
        \delta \bm{\theta}_A &\gets \mathbf{R}^T\left\{\bm{w}_{mA}\Delta t \right\} \delta \bm{\theta}_A -\bm{w}_{nA} \Delta t \tag{\ref{delta_x}c} \\
        \delta \bm{\theta}_B &\gets \mathbf{R}^T\left\{\bm{w}_{mB}\Delta t \right\} \delta \bm{\theta}_B - \bm{w}_{nB} \Delta t \tag{\ref{delta_x}d} \label{delta_x_d}
    \end{align}
    where $\bm{\alpha}=-\mathbf{R} \left\{ \mathbf{q} \right\} [\mathbf{a}_{mA}]_{\times} \delta \bm{\theta}_A + [\mathbf{a}_{mB}]_{\times} \delta \bm{\theta}_B - \mathbf{R} \left\{ \mathbf{q} \right\} \mathbf{a}_{nA }+ \mathbf{a}_{nB}$ and the definition of the cross-product matrices $[\enspace]_{\times}$ can be found in \cite{sola2017quaternion}. 
    
     We define $\mathbf{F_x}$ and $\mathbf{F_i}$ are the Jacobians of $f$ with respect to $\delta \mathbf{x}$ and $\mathbf{u}_n$, $\mathbf{F_x}$ and $\mathbf{F_i}$ are calculated by
    \begin{equation}
        \mathbf{F_x} = {\frac{\partial f}{\partial \delta \mathbf{x}}} \Bigg|_{\mathbf{x},\mathbf{u}_m} ,\quad
        \mathbf{F_i} = {\frac{\partial f}{\partial \delta \mathbf{u}_n}} \Bigg|_{\mathbf{x},\mathbf{u}_m} 
    \end{equation}
    
    Then the prediction equations can be written as
    \begin{align}
        \hat{\delta \mathbf{x}} &\gets \mathbf{F_x}(\mathbf{x},\mathbf{u}_m) \hat{\delta \mathbf{x}} \label{delta_x_update} \\
        \mathbf{P} &\gets \mathbf{F_x} \mathbf{P} \mathbf{F_x}^T + \mathbf{F_i} \mathbf{Q_i} \mathbf{F_i}^T \label{P_update}
    \end{align}
    where ${\delta \mathbf{x}}\sim{\mathcal{N} ( \hat{\delta \mathbf{x}}, \mathbf{P})}$ and $\mathbf{Q_i}$ is the covariance matrix of $\mathbf{u}_n$.
    
    When IMU data is received, we follow equations (\ref{nominal_state_start}) $\sim$ (\ref{nominal_state_end}), equations (\ref{delta_x_update}) and (\ref{P_update}) to update the nominal-state, error-state and error-state covariance matrix, respectively.

    \subsubsection{Measurement Model}
    We take the raw calculation results in section \ref{sec:direct computation} as measurements $\mathbf{z}$, 
    \begin{equation}
        \mathbf{z}=\left[\begin{array}{c}
                {^B\widetilde{\mathbf{p}}}_A\\
                {^A\widetilde{\mathbf{p}}}_B\\
                ^B{\widetilde{\mathbf{q}}}_A\\
            \end{array}\right]
    \end{equation}
    where $^B{\widetilde{\mathbf{q}}}_A$ corresponds to $^B{\widetilde{\mathbf{R}}_A}$. Since the same sensors' observations are used to compute $^B{\widetilde{\mathbf{R}}_A}$ and $^A{\widetilde{\mathbf{R}}_B}$, we only need to select one of them as the measurements. 
    The relationship between the measurements $\mathbf z$ and the true-state is written as
    \begin{align}
        \mathbf{z} &= h(\mathbf{x}_t)+\bm{v} \label{z}  \\
        {^B\widetilde{\mathbf{p}}}_A &= \mathbf{p}_t + \mathbf{p}_{nA} \tag{\ref{z}{a}} \\
        {^A\widetilde{\mathbf{p}}}_B &= -\mathbf{R}^T\left\{ \mathbf{q}_t \right\} \mathbf{p}_t + \mathbf{p}_{nB} \tag{\ref{z}{b}} \\
        ^B{\widetilde{\mathbf{q}}}_A &= \mathbf{q}_t + \mathbf{q}_{n} \tag{\ref{z}{c}}
    \end{align}
    where $\bm{v} = {[\mathbf{p}_{nA}, \mathbf{p}_{nB}, \mathbf{q}_{n}]}^T \sim{\mathcal{N} (0, \mathbf{V}})$ is a white Gaussian noise with the covariance V.
    
    The true state estimation can be calculated by $\hat{\mathbf{x}}_t=\mathbf{x} \oplus \hat{\delta \mathbf{x}}$. As the error-state mean $ \hat{\delta \mathbf{x}} = 0$, we have $\hat{\mathbf{x}}_t=\mathbf{x} $. Therefore, we take $\mathbf{x}$ as the evaluation point and the Jacobian matrix of the measurement model $\mathbf{H}$ is
    \begin{equation}
        \mathbf{H}={\frac{\partial h}{\partial \delta \mathbf{x}}} \Bigg|_{\mathbf{x}} ={\frac{\partial h}{\partial \mathbf{x}_t}} \Bigg|_{\mathbf{x}} {\frac{\partial \mathbf{x}_t}{\partial \delta \mathbf{x}}} \Bigg|_{\mathbf{x}}
    \end{equation}
    
    The correction equations can be written as
    \begin{align}
        \mathbf{K} &= \mathbf{P}\mathbf{H}^T{(\mathbf{HPH}^T + \mathbf{V})}^{-1} \\
        \hat{\delta \mathbf{x}} &\gets \mathbf{K}(\mathbf{z}-h(\hat{\mathbf{x}_t})) \label{delta_x_observe}\\
       \mathbf{P} &\gets \mathbf{(I-KH)P} \label{P_observe}
    \end{align}
    
    We use equation (\ref{delta_x_observe}) and equation (\ref{P_observe}) to compute the observed error and update the error-state covariance matrix, respectively.
    
    \subsubsection{Error Injection and Reset}
    When the measurements update is finished, we add the observed error to the nominal state by
    \begin{equation}
        \mathbf{x} \gets \mathbf{x} \oplus \hat{\delta \mathbf{x}} \label{nominal_state_update}
    \end{equation}

    After the error injection step, we reset the error state for the next iteration by
    \begin{align}
        \delta \mathbf{x} & \gets g(\delta \mathbf{x}) = \delta \mathbf{x} \ominus \hat{\delta \mathbf{x}} \label{error_state_reset}\\
        \delta \mathbf{p} & \gets \mathbf{R}^T \left\{\hat{{\delta \bm{\theta}}_B}\right\}(\delta \mathbf{p} - \hat{\delta \mathbf{p}}) \tag{\ref{error_state_reset}a} \label{error_reset_a}\\
        \delta \mathbf{v} & \gets \mathbf{R}^T \left\{\hat{{\delta \bm{\theta}}_B}\right\}(\delta \mathbf{v} - \hat{\delta \mathbf{v}}) \tag{\ref{error_state_reset}b} \label{error_reset_b}\\
        {\delta \bm{\theta} }_A &\gets - \hat{{\delta \bm{\theta} }_A} + \left(\mathbf{I}-{\left[\frac{1}{2}\hat{{\delta \bm{\theta} }_A}\right]}_{\times}\right) {\delta \bm{\theta}}_A \tag{\ref{error_state_reset}c} \label{error_reset_c}\\
        {\delta \bm{\theta} }_B &\gets - \hat{{\delta \bm{\theta} }_B} + \left(\mathbf{I}-{\left[\frac{1}{2}\hat{{\delta \bm{\theta} }_B}\right]}_{\times}\right) {\delta \bm{\theta}}_B \tag{\ref{error_state_reset}d} \label{error_reset_d}
    \end{align}
    
    We define $\mathbf{G}$ as the jacobian matrix of the error reset function (\ref{error_state_reset}), $\mathbf{G}$ can be computed by
    \begin{equation}
        \mathbf{G} = {\frac{\partial g}{\partial \delta \mathbf{x}}} \Bigg|_{\hat{\delta \mathbf{x}}}
    \end{equation}
    
    Finally, we update the error-state mean $\hat{\delta \mathbf{x}}$ and its covariance matrix $\mathbf{P}$ by
    \begin{align}
        \hat{\delta \mathbf{x}} &\gets \mathbf{0} \\
        \mathbf{P} &\gets \mathbf{GPG}^T
    \end{align}
   \subsection{Pose Graph Optimization}

    \begin{figure}[t]
   	\vspace{0.2cm}
   	\centering
   	\includegraphics[width=0.9\linewidth]{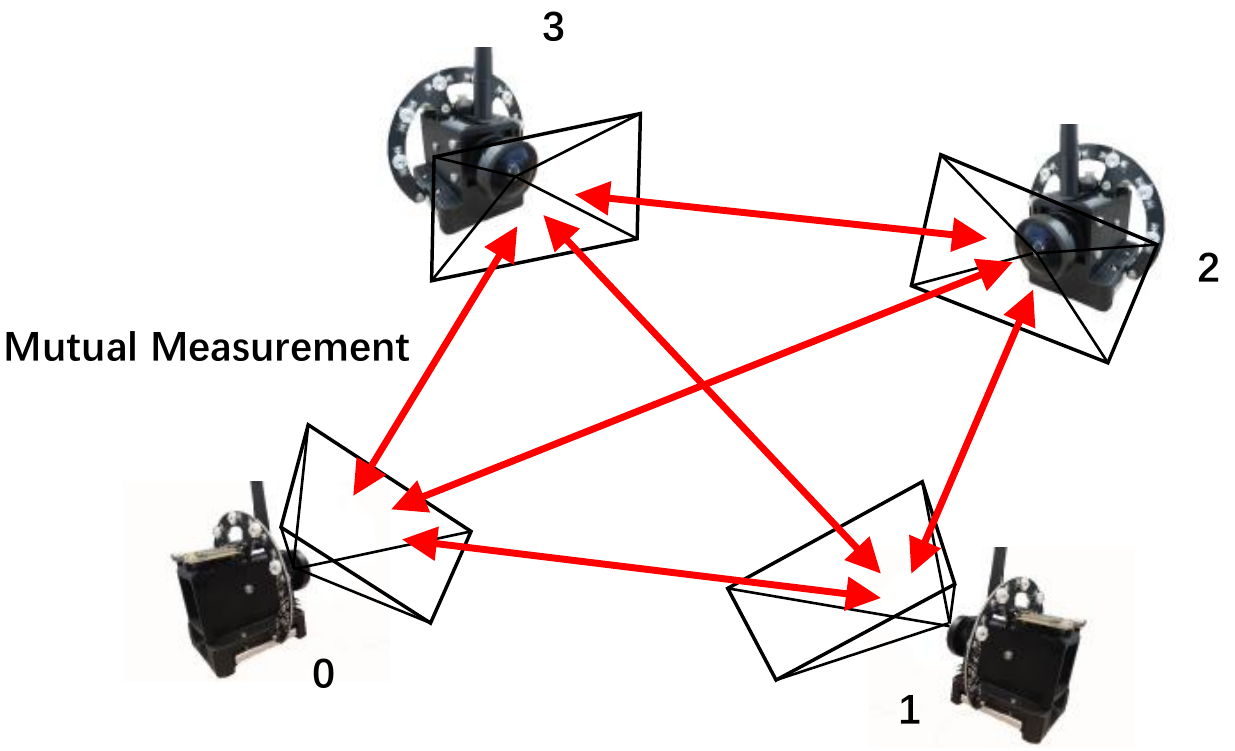}
   	\captionsetup{font={small}}
   	\caption{
   		An illustration of PGO of 4 robots
   	}
   	\label{pic:pgo}
   	\vspace{-1.4cm}
   \end{figure}
   
   From the above sections, we get the refined relative pose estimation between two robots. When multiple robots are around, we propose a PGO-based algorithm to further improve the mutual localization accuracy. Unlike the classic PGO using multi-frame measurements in the continuous time domain, our PGO formulation is for any single frame of mutual measurements. As shown in Fig. \ref{pic:pgo}, each robot represents a node in the graph and the edge is the mutual relative pose between two robots.
   %any three robots can form the smallest interloop. Inside an interloop, such as robots 1, 2, and 3 in Fig. \ref{pic:pgo}, each edge has more than two independent measurements. As the number of robots increases, so does the number of interloops in the system. The graph include all available interloops inside a MRS.
   
   Currently, each robot runs the PGO algorithm in a distributed manner after receiving all the available mutually measured poses from the neighbors. For an arbitrary robot, we denote its coordinate frame as ${C}$, the pose of robot $i$ in $C$ as ${\mathbf X}_i = ({\mathbf R}_i, {\mathbf t}_i) \in SE(3)$, the measured relative pose between robot $i$ and robot $j$ as ${\hat{\mathbf T}}_{i j} = (\hat{{\mathbf R}}_{i j}, \hat{{\mathbf t}}_{i j}) \in SE(3), i \neq j$, and the PGO problem can be formulated as follows
   \begin{equation}
   \min _{\mathrm{{\mathbf X} \in \mathbf O}} \sum_{(i, j) \in \mathbf L} \rho (r_{i j}({\mathbf X}_i, {\mathbf X}_j, \hat{\mathbf T}_{i j}))\label{pgo_equation}
   \end{equation}
   where ${\mathbf O}$ is the set of robots, ${\mathbf L}$ is the set of robot couples, $\rho( )$ is the kernel function, and $r_{i j}({\mathbf X}_i, {\mathbf X}_j, \hat{\mathbf T}_{i j})$ is defined as
   
   \begin{equation}
   r_{i j}({\mathbf X}_i, {\mathbf X}_j, \hat{\mathbf T}_{i j}) = \Vert \hat{{\mathbf T}}_{i j} \cdot ({{\mathbf X}}_{j}^{-1} \cdot {{\mathbf X}}_{i}) -{\mathbf{I}}\Vert_F^2
   \end{equation}
   	
   We use open-sourced GTSAM \cite{gtsam} to solve the graph optimization.

   \section{EXPERIMENT}
    
   To show the accuracy and features of our system, we design a series of experiments with UAVs and Unmanned Ground Vehicles (UGVs). MCS and RTK are introduced as ground truth. We use the error evaluation method in \cite{zhang2018tutorial} to demonstrate the mutual localization accuracy of our system. Our experiments contain two parts: 1). Accuracy comparison in two-robot and multi-robot scenarios. 2). Feature validation experiments. Considering the computation time firstly, the ID extraction takes less than 2 ms per image, raw relative pose estimation and ESKF iteration takes less than 1 ms, and the PGO needs around 3 ms. The selected computation platform, Intel NUC-i5, is proven to have sufficient computation resources.
   
   \subsection{Accuracy Comparison}
   \subsubsection{Two-Robot Mutual Localization}
   	\begin{figure*}[ht]
   	\centering
   	\captionsetup{font={small}}
   	\includegraphics[width=1.0\linewidth]{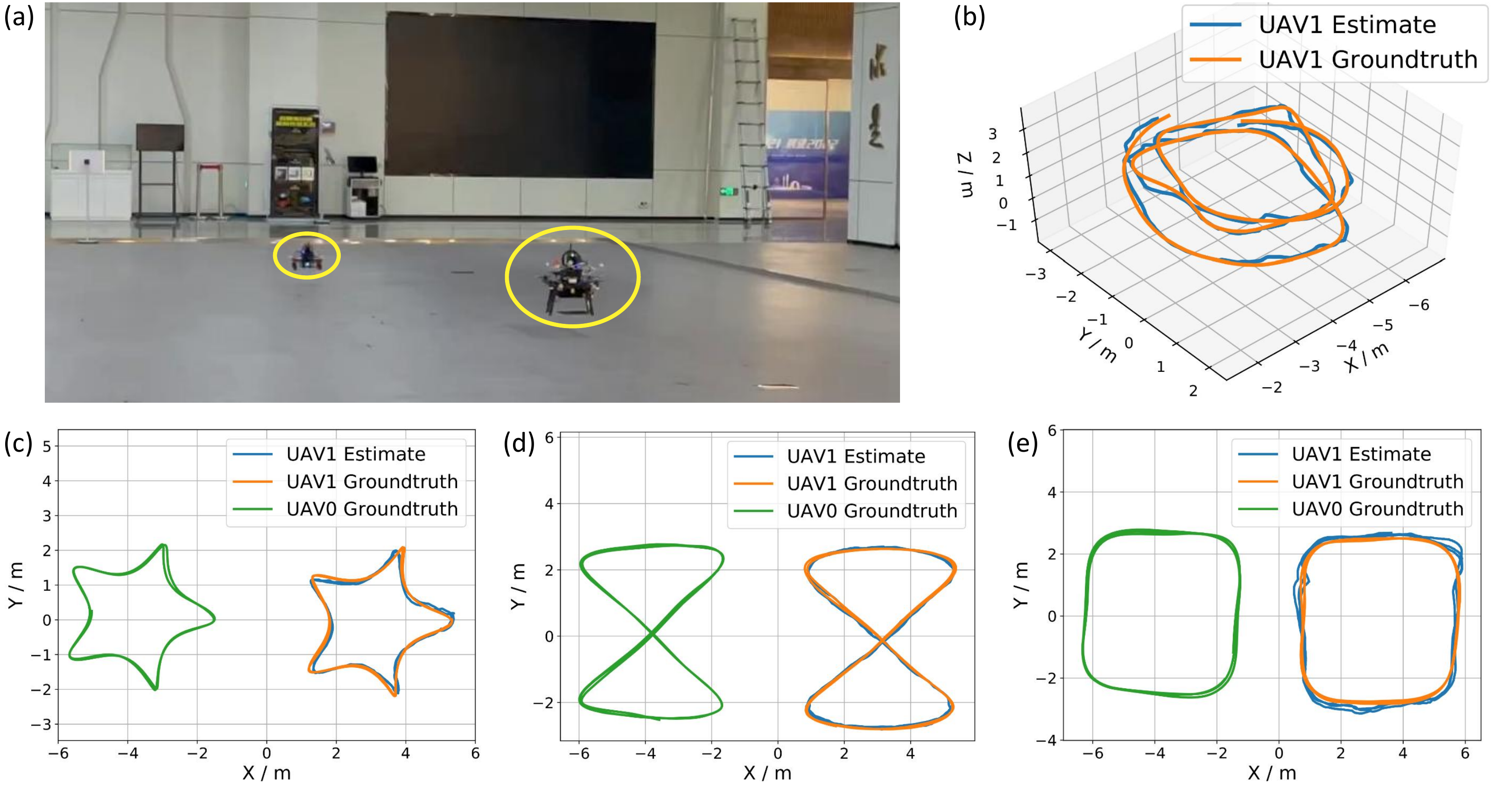}
   	\captionsetup{font={small}}
   	\caption{
   		Two robots mutual localization. (a) shows our experimental scenario. (b) shows trajectories in 3D while UAVs are in manual control. Trajectories in (c)(d)(e) are shown in the top view while UAVs are programmed to fly autonomously.
   	}
   	\label{pic:2robotexp}
   	\vspace{-0.2cm}
   \end{figure*}
   
   \begin{figure}[t]
   	\centering
   	\captionsetup{font={small}}
   	\includegraphics[width=0.98\linewidth]{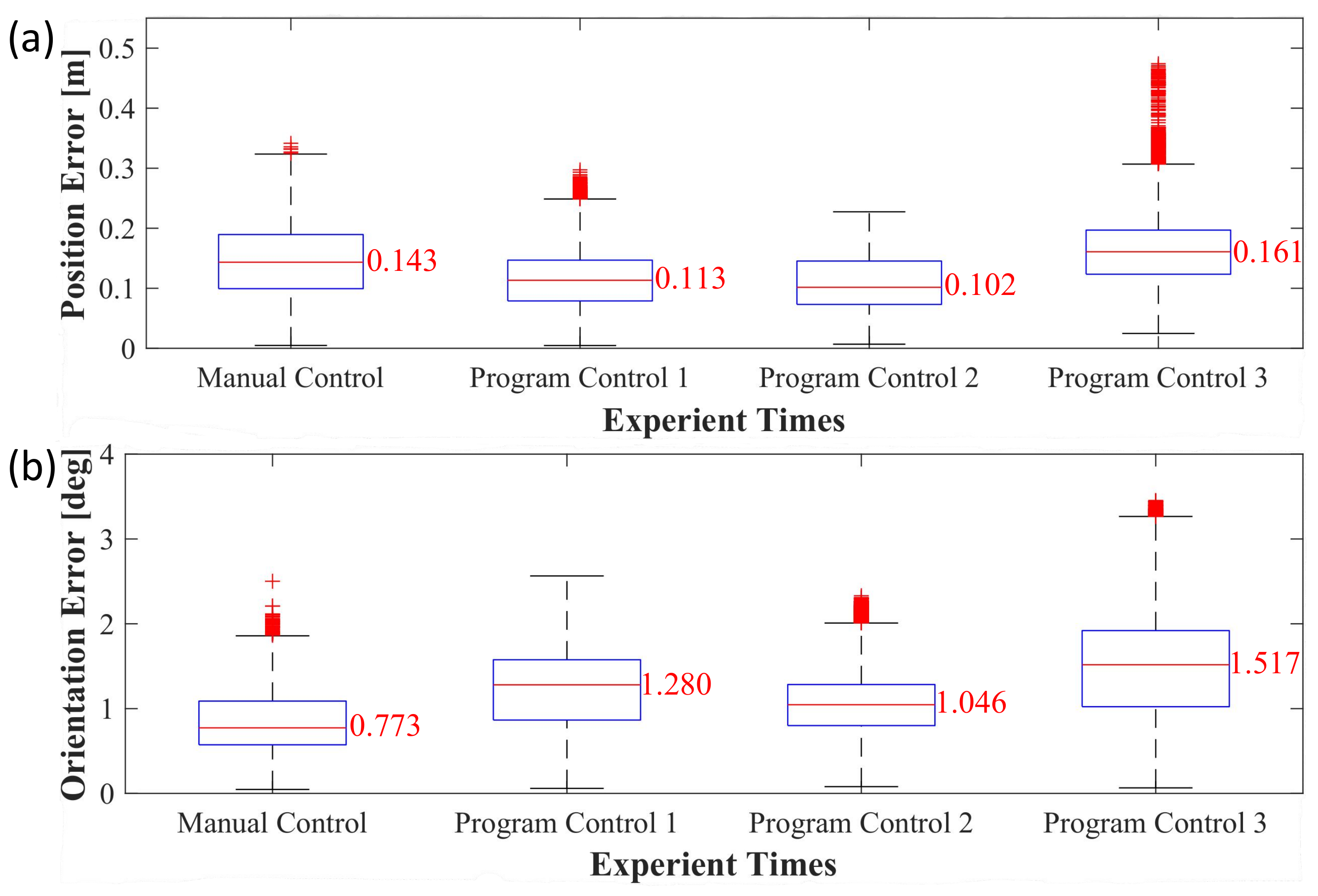}
   	\caption{
   		Boxplot of the position error (a) and orientation error (b) for experiments shown in Fig \ref{pic:4robotexp}.
   	}
   	\label{pic:2robotexp_accuracy}
   	\vspace{-1.5cm}
   \end{figure}

   Experiments are implemented indoors, and MCS is used as ground truth. As shown in Fig.\ref{pic:2robotexp}, to demonstrate the robustness of our system, we conduct one manual control and multiple autonomous control experiments with two UAVs.
   To effectively visualize the comparison, we calculate the estimated trajectory of UAV1 by adding the relative pose estimations on the ground truth of UAV0, the observer.
   Experiments show that our system can consistently and stably provide relative pose estimations. We can see from Fig.\ref{pic:2robotexp_accuracy} that our system achieves high relative pose estimation accuracy. For all experiments, the median of mutual position estimation errors is in the range of 0.102 to 0.161 meters, and the median of mutual orientation errors is in the range of 0.733 to 1.517 degrees.
   
   \subsubsection{Multi-Robot Mutual Localization}

	\begin{table}[!htbp] 
		\centering
		\caption{Accuracy comparison of four-robot mutual localization experiment in UAV0 body frame}
		\begin{tabular}{cccc}
			\toprule 
			& without  & with & Improvement \\
			& PGO   &  PGO & \\ 
			\hline
			Traj. Lengths of UAV1, UGV0,1 & \multicolumn{3}{c}{ {(22.0, 17.7, 11.6)} m } \\ 
			\hline
			$\sum$ $ATE_{pos}$($\mathbf{X}_{i}^0$)$/n$   & 0.089m   &0.073m   & 0.016m \\  
			\hline
			$\sum$ $ATE_{rot}$($\mathbf{X}_{i}^0$)$/n$   & 0.884$^{\circ}$   &0.879$^{\circ}$   & 0.005$^{\circ}$ \\ 
			\bottomrule
		\end{tabular}
		\label{tab:performance_pgo}
		\vspace{-0.9cm}
	\end{table}
   
   \begin{figure}[h]
   	\centering
   	\captionsetup{font={small}}
   	\includegraphics[width=0.9\linewidth]{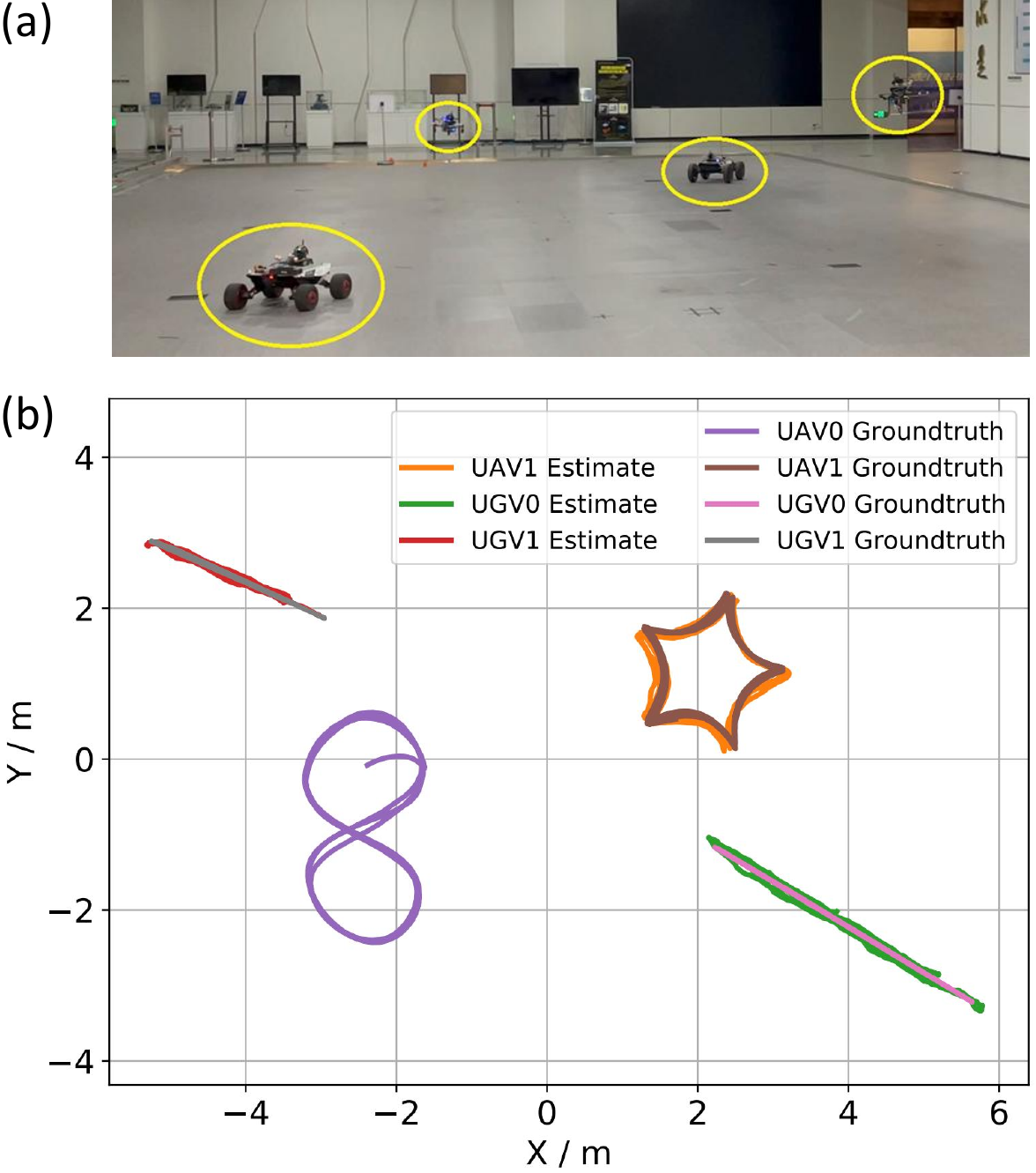}
   	\caption{Multi-robot mutual localization}
   	\label{pic:4robotexp}
   	\vspace{-0.22cm}
   \end{figure}
   
   Four robots, two UGVs and two UAVs, are prepared to prove the multi-robot mutual localization accuracy, as shown in Fig.\ref{pic:4robotexp} (a). Different from the above experiments, we add PGO to improve multi-robot mutual localization performance. Fig.\ref{pic:4robotexp} (b) shows the estimated trajectories of other robots with respect to the UAV0 frame (as the observer). Table \ref{tab:performance_pgo} shows the average Absolute Trajectory Error (ATE) \cite{zhang2018tutorial} for $n$ robots in the UAV0 body frame, under the condition of with and without PGO. Results show that PGO improves the accuracy of position and orientation estimation, albeit by a small amount in the clear condition. 
   % However, the PGO has another important function is to collaborative estimation for robots that are not under line-of-sight conditions, testified in the following section.

   \subsection{Feature Validation}
   
   \subsubsection{Dark Scenario}
   
   \begin{figure}[h]
   	\centering
   	\includegraphics[width=1.0\linewidth]{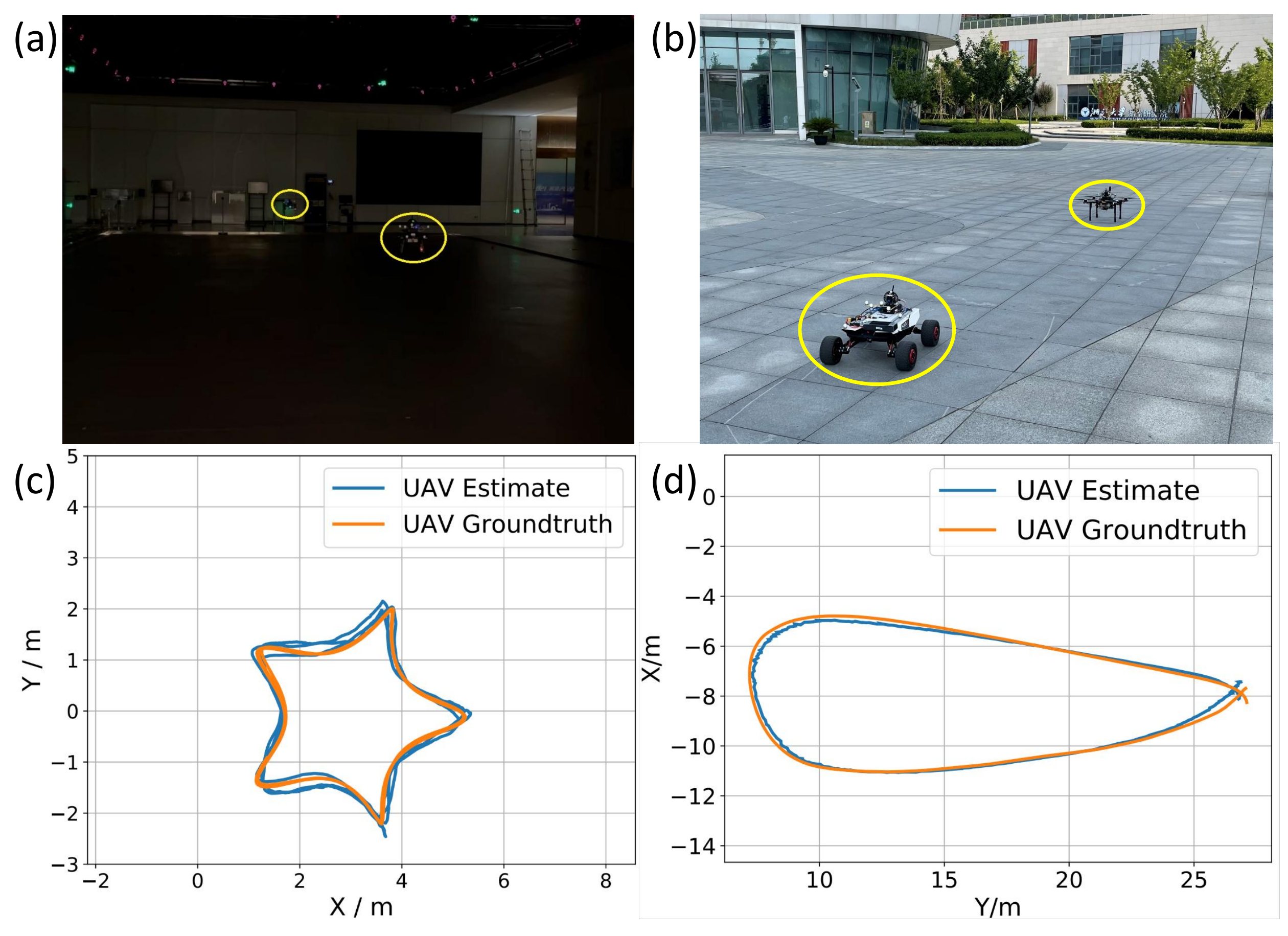}
   	\captionsetup{font={small}}
   	\caption{
   		Dark and Long-Range Scenarios Experiments.  
   	}
   	\label{pic:feature}
   	\vspace{-1.05cm}
   \end{figure}
   
   As shown in Fig.\ref{pic:feature} (a), we test the system in an almost totally dark environment, which is challenging for VIO-based indirect methods. Similar to the two-UAV experiments, the observer can consistently estimate the relative pose of the other UAV as shown in Fig.\ref{pic:feature} (c). This experiment verifies that the proposed system can work in environments with low-light conditions.
   
   \subsubsection{Long-Range Scenario}
   
    As shown in Fig.\ref{pic:feature} (b), we conduct the long-range experiment outdoors with a UAV and a UGV. During the experiment, the UGV is static, and the UAV is flying under manual control. Since MCS can't be deployed in the outdoor environment, we use RTK GPS as our ground truth for comparison. Fig.\ref{pic:feature} (d) shows the system can stably estimate peer poses far from 27.5 meters, which shows better support in large areas than works \cite{8967660}\cite{faessler2014monocular,cutler2013lightweight,dias2016board}  using active LEDs (relative pose estimated within 6 meters distances). 
    
    \subsubsection{Aggressive Motion Scenario}
    	\begin{figure}[t]
    		\centering
    		\captionsetup{font={small}}
    		\includegraphics[width=1.00\linewidth,height=0.65\linewidth]{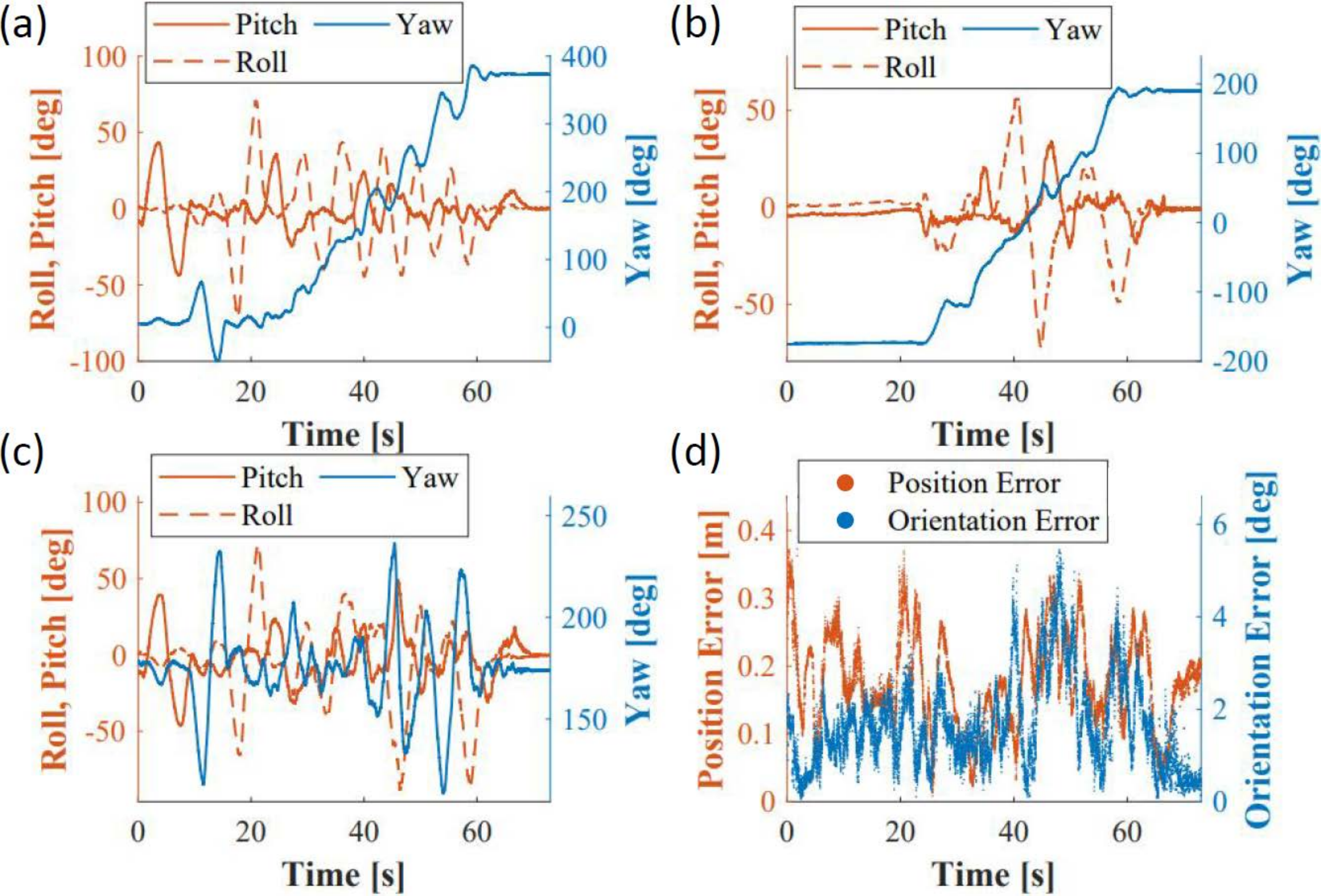}
    		\caption{Aggressive Motion Experiment. Sub-figure (a),(b) show the ground truth orientation changes of robot0 and robot1. (c) shows the ground truth relative pose between the two robots. From (a),(b), and (c), we can see aggressive motion between robot0 and robot1. Then (d) shows the relative pose estimation errors.}
    		\label{pic:bigposetexp}
    		\vspace{-0.9cm}
    	\end{figure}
    	To testify our system under extreme conditions, we perform an aggressive motion experiment with large roll/pitch/yaw angular changes. The experiment is conducted via two handheld devices as it is difficult to control UAVs/UGVs to perform such large rotation angular excursions, either manually or autonomously. Two people walk around a circle (with a diameter of 4 meters) and move the two devices' attitudes randomly. As the absolute poses shown in Fig. \ref{pic:bigposetexp}(a) and Fig. \ref{pic:bigposetexp}(b), the maximum changing range for pitch angle reaches 87 degrees, for roll angle reaches 138 degrees and for yaw reaches 360 degrees. The relative poses also vary over large angles as shown in Fig. \ref{pic:bigposetexp}(c). From Fig. \ref{pic:bigposetexp}(d), we see the relative position error is below 0.4 meters with a median of 0.174 meters and the angle error is below 6 degrees with a median of 1.48 degrees.

  \subsubsection{Cooperative Localization in Occluded Scenario}
  
  \begin{figure}[h]
  	\centering
  	\includegraphics[width=1.0\linewidth]{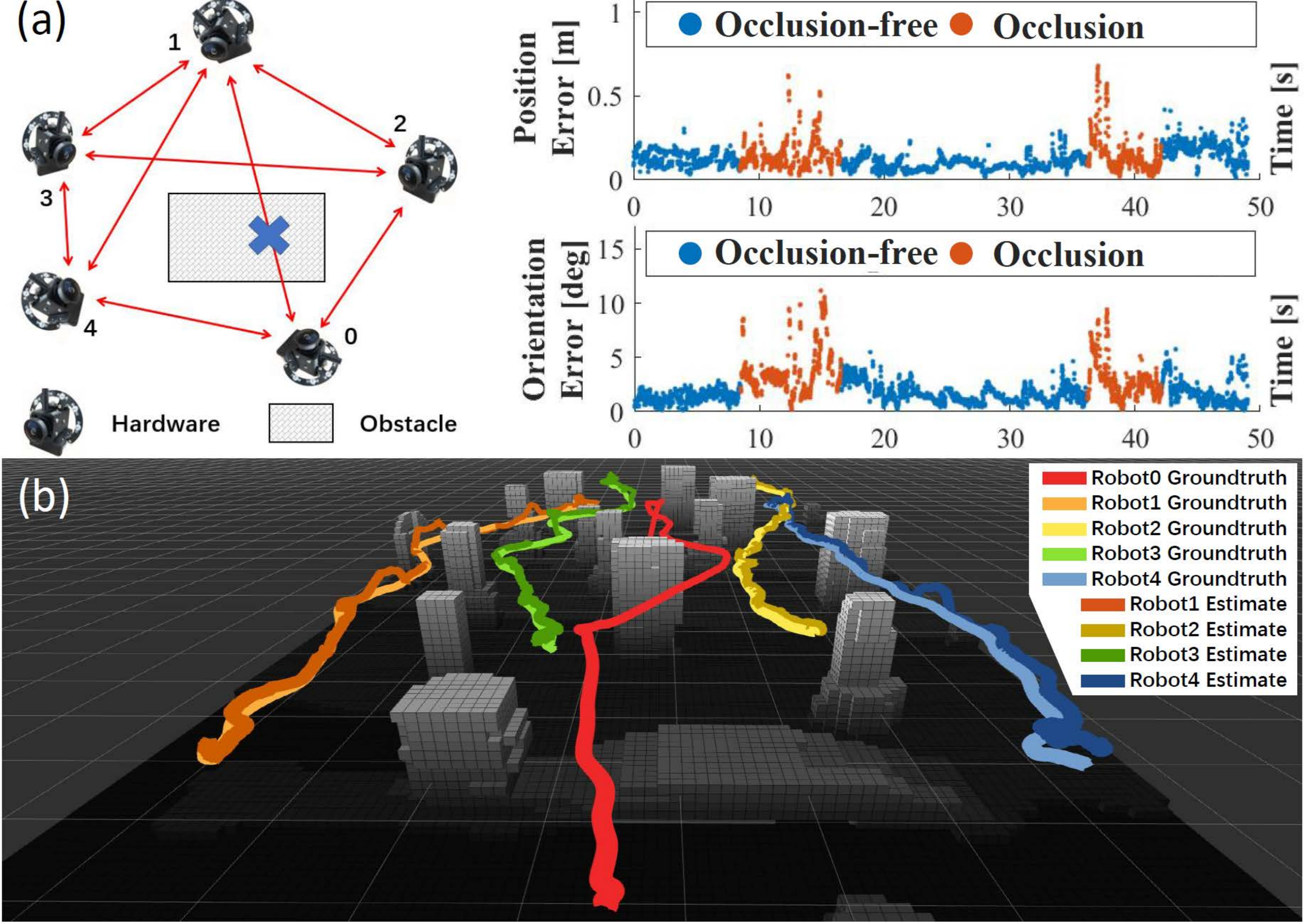}
  	\captionsetup{font={small}}
  	\caption{
	Occlusion experiment. 
 % (a) shows a pose graph of five robots under a wall-like occlusion condition and the position and orientation errors of robot1 in robot0 frame respectively. 
 We conduct experiments with five robots in two scenarios with different levels of occlusion.
 (a) shows the position and orientation errors of robot1 in robot0 frame under an isolated obstacle occlusion condition.
 (b) shows the remapped trajectories of robot0'neighbors into its own frame in a complex environment, compared with neighbors' ground truth trajectories.
  	}
  	\label{pic:nlos}
  	\vspace{-0.15cm}
  \end{figure}
	
	Occlusion can not be avoided when deploying robot teams in real-world applications, which could fail the relative pose estimation between two robots as the lack of mutual visual measurements. In this situation, the PGO-based algorithm could be used to recover the relative poses through cooperative localization in certain conditions.
 % We perform a five-robot experiment in a highly cluttered environment to demonstrate NLOS performance. 
	 Firstly we compare the accuracy under occluded and non-occluded conditions.
	 We move a group of five robots in an environment with an isolated obstacle, where robot0 and robot1 happen to be occluded by the obstacle at some points as shown in the left of Fig.\ref{pic:nlos} (a). The right of Fig.\ref{pic:nlos} (a) shows the estimation error and we can see the error is slightly larger when occlusion happens, which is acceptable considering there’s no direct measurements between robot1 and robot0. Secondly, we conduct the other experiment in a much more complex environment with many obstacles as shown in Fig.\ref{pic:nlos} (b). We can see the estimated trajectories match well with ground truth trajectories. 
   
   \section{Conclusion}
   \label{sec:conclusion}
   This paper introduces CREPES, a novel, robust and accurate solution for multi-robot mutual localization. We have conducted extensive experiments to show the performance of CREPES in two-robot and multi-robot situations, even in dark or large-scale environments, with aggressive motion or under occlusion conditions. Our relative pose estimation system can achieve a median of 0.13 meters position accuracy and a median of 1.16 degrees orientation accuracy in clear conditions.
     
   Although we testify our system under occlusion conditions and show similar accuracy, we find there are much more outliers than in clear conditions. 
   % In the future, we will consider tackling the occluded scenarios through multi-robot cooperative optimization in the continuous time domain. 
   % In future research, we will target the challenge of multi-robot cooperative optimization in the continuous time domain. 
   In future research, we will tackle this challenge from the perspective of multi-robot cooperative optimization in continuous time domain.
   In addition, we will iterate the hardware to make the system smaller for robotic applications.
   % In addition, we will keep improving the hardware setup and try to make the physical attribute fit robotic applications.
    
    \newlength{\bibitemsep}\setlength{\bibitemsep}{0.0\baselineskip}
	\newlength{\bibparskip}\setlength{\bibparskip}{0pt}
	\let\oldthebibliography\thebibliography
	\renewcommand\thebibliography[1]{%
		\oldthebibliography{#1}%
		\setlength{\parskip}{\bibitemsep}%
		\setlength{\itemsep}{\bibparskip}%
	}
	\bibliography{RAL}

\end{document}